\documentclass[]{spie}  
\usepackage[]{graphicx}
\usepackage[colorlinks=true, allcolors=blue]{hyperref}
\title{Hyperspectral Pigment Analysis of Cultural Heritage Artifacts Using the Opaque Form of Kubelka-Munk Theory} 

\author{Abu Md Niamul Taufique} 
\author{David W. Messinger}
\affil{Rochester Institute of Technology, Rochester, New York}

\authorinfo{CONTACT Abu Md Niamul Taufique, Email: at7133@rit.edu}

\begin{document} 
  \maketitle 

\begin{abstract}
Kubelka-Munk (K-M) theory has been successfully used to estimate pigment concentrations in the pigment mixtures of modern paintings in spectral imagery. In this study the single-constant K-M theory has been utilized for the classification of green pigments in the Selden Map of China, a navigational map of the South China Sea likely created in the early seventeenth century.  Hyperspectral data of the map was collected at the Bodleian Library, University of Oxford, and can be used to estimate the pigment diversity, and spatial distribution, within the map. This work seeks to assess the utility of analyzing the data in the K/S space from Kubelka-Munk theory, as opposed to the traditional reflectance domain.  We estimate the dimensionality of the data and extract endmembers in the reflectance domain.  Then we perform linear unmixing to estimate abundances in the K/S space, and following Bai, et al. (2017), we perform a classification in the abundance space.  Finally, due to the lack of ground truth labels, the classification accuracy was estimated by computing the mean spectrum of each class as the representative signature of that class, and calculating the root mean squared error with all the pixels in that class to create a spatial representation of the error. This highlights both the magnitude of, and any spatial pattern in, the errors, indicating if a particular pigment is not well modeled in this approach.  
\end{abstract}


\section{Introduction}
The Selden Map of China was an early seventeenth century map which was brought into the Bodleian Library at the University of Oxford in 1659 \cite{bai2018, batchelor2013, Kogou2016}. The map was named after John Selden, a renowned London lawyer who donated it to the Bodleian Library in 1654 \cite{Kogou2016}. This map provides significant understanding of the globalization in Asia in the early seventeenth century. In 2008, the map was `rediscoverd' by Robert Batchelor \cite{batchelor2013} which give researchers an opportunity to understand Chinese cartography, the timeframe of the creation of the map, material diversity of the map, merchant shipping history in Asia and disputes about the merchant routes of the Ming Empire \cite{batchelor2013}. Though there is a dispute if the map was legally obtained or not, it is the first Chinese map of merchant data that reached England and the first of its kind to survive \cite{batchelor2013}. The map was painted on paper with black carbon ink and six different colors: red, green, yellow, blue, white, and black portraying flowers, trees, rivers, and mountains \cite{batchelor2013}. To understand the pigment material diversity of the map, hyperspectral image (HSI) data of the map were collected at the Bodleian Library at Oxford University in 2015. To get high spatial resolution image, the overall collection of the HSI data was split into 12 `chips'. The spatial size of each chip may vary from 1600 (pixels) x 2300 (pixels) to 1600 (pixels) x 3250 (pixels). Each HSI chip contains the reflectance measurement in the range of visible to near infra-red, i.e., 224 channels from 400nm to 1000nm. In section \ref{sec:experiment} we will discuss the details of the data collection parameters. 

Among many, two common studies on cultural heritage artifacts are faded text enhancement and pigment analysis which play significant role in the codicological studies (study of codices or manuscripts written on parchment or paper). For faded text analysis there are several methods which are based on multispectral imaging techniques such as reflectance, fluoresence, and transmission based techniques \cite{easton2011ten, easton2010infinite, christens2011some, easton2011some, knox2011recovery, easton2014statistical, easton2003multispectral, easton2015rediscovering}. Techniques that have been used in these analysis are mainly principal component analysis (PCA), independent component analysis (ICA) and different supervised or unsupervised clustering methods \cite{easton2011ten, easton2010infinite, christens2011some, easton2011some, knox2011recovery, easton2014statistical, easton2003multispectral, easton2015rediscovering}. 

Hyperspectral imaging techniques have become more widely used for historical artifact analysis in recent years. Bai et al \cite{bai2018, bai2017hyperspectral, bai2017pigment} has done extensive works on HSI data classification of historical artifacts. In these works most of the techniques involve unmixing which is adopted from the remote sensing community. For example, Bai et al \cite{bai2017pigment} used HSI data to estimate the within pigment material diversity of Gough Map data where they used some initial processing of the HSI data by sphering and binning the data matrix. Then they used the Gram matrix and MaxD techniques to estimate the dimensionality of the data and Spectral Angle Mapper (SAM) to estimate the final classification map. 

Unmixing the mixture of pigments to find the constituent pigments is inherently different from the linear unmixing techniques used in the field of remote sensing \cite{bowles2007optical, zhang2005spectral}. In remote sensing imagery, the spatial area sampled by each pixel is much larger and the measured radiance in each pixel is the effect of different materials present within this area. Here the materials are spatially separated in most of the cases. As a result, linear unmixing theory in the reflectance space is sufficient to unmix the spectra of the constituent materials. In contrast, the paint mixtures usually have small particles uniformly distributed in the binding medium which we call intimate mixtures. So the total reflectance is not just a linear combination of the constituent pigments' spectral reflectance in the reflectance space \cite{liang2012advances}. Rather, the spectral unmixing for pigment mixtures are best modeled by a physics based model of light transport in turbid medium.  

The Kubelka-Munk (K-M) theory \cite{kubelka1931contribution, kubelka1948new} proposed a radiative transfer model involving two diffused fluxes that propagates in the forward and backward direction. Because of the simplicity and accuracy of the model in paint unmixing (finding the mixing ratios of paints to match a given color), this method is still widely used in paint industries. Significant work has been done in the field of pigment identification using K-M theory \cite{zhao2008investigation, moghareh2017linear, liang2008pigment} with multispectral image (MSI) data. These works mainly focus on pigment identification using an existing database of unique pigments. For instance, Zhao et al \cite{zhao2008investigation} extensively studied Vincent van Gogh's \textit{The Starry Night} using single-constant K-M theory where the author used preexisting database of pure pigments to find the best match. However, there is still a lack of study for a pigment diversity estimation using the HSI data without any prior database of unique pigments.

In this paper we implemented the single-constant K-M theory for the pigment material diversity estimation without using any prior database of unique pigments. Our results show that single-constant K-M theory is useful in pigment mapping in the field of historical document image analysis even if we have a little or no knowledge about the pigments present. The rest of the paper is organized as follows.
In section \ref{sec:K-Mtheory} we discuss the theoretical background of K-M theory. 
In section \ref{sec:pigmentdiversity} we describe the dimensionality estimation and unmixing of the HSI data. 
In section \ref{sec:method} we present the methodology of our experiment.
In section \ref{sec:experiment} we demonstrate the application of the method on the Selden Map.
In section \ref{sec:summary} we summarize the results and draw conclusions.  

\section{Kubelka-Munk Theory}
\label{sec:K-Mtheory}
The K-M theory \cite{kubelka1931contribution, kubelka1948new} describes the radiative transfer model of two diffuse fluxes in a turbid medium. The diffuse reflectance $R$ of a film relates to the effective absorption coefficient $K$ and effective scattering coefficient $S$ using the following formulae
\begin{equation}
   R = \frac{1 - R_g [a - bCoth(bSh) ]}{a + bCoth(bSh) - R_g} 
\end{equation}
where $R_g$ is the reflectance of the background, $h$ is the film thickness, $a = (K+S)/S$, and $b = \sqrt{a^2 -1}$. Considering an opaque layer, the reflectance factor $R_{\infty}$ can be written as
\begin{equation}
    R_{\infty} = 1 + \frac{K}{S} - \sqrt{\left( \frac{K}{S} \right)^2 + 2\left( \frac{K}{S} \right)}
\end{equation}
This can be written as
\begin{equation}\label{eq:eq3}
    \frac{K}{S} = \frac{\left(1 - R_{\infty} \right)^2}{2 R_{\infty}}
\end{equation}

Using the assumption of additivity and scalibility \cite{duncan1940colour}, the absorption and scattering coefficients of the paint mixtures are the linear combination of the absorption and scattering coefficients of the component paints weighted by their concentrations, respectively. This is referred to as a two-constant K-M theory. However, when the scattering is dominated by white pigment, the ratio, K/S of the overall pigment mixture can be modeled as the linear combination of the K/S of all the constituent pigments \cite{berns2002multiple}. This is referred to as the single constant K-M theory. We can write the relationship for the single constant K-M theory as \cite{guo1998application}  

\begin{equation}\label{eq:eq4}
    \left( \frac{K}{S} \right)_{MixOnPaper} = \left( \frac{K}{S} \right)_{Paper} + c_{1}\left( \frac{K}{S} \right)_{Pigment-1} + c_{2}\left( \frac{K}{S} \right)_{Pigment-2} + .... + c_{n}\left( \frac{K}{S} \right)_{Pigment-n}
\end{equation}
Where, 
\begin{equation}
    \left( \frac{K}{S} \right)_{Pigment-n} = \left( \frac{K}{S}\right)_{PigmentOnPaper -n} - \left( \frac{K}{S} \right)_{Paper}
\end{equation}
Here, MixOnPaper and PigmentOnPaper represent the spectral properties of the paintings and masstone of the pigments on paper, respectively. Replacing this value in equation \ref{eq:eq4} we get 
\begin{equation}\label{eq:eq6}
    \left( \frac{K}{S} \right)_{MixOnPaper} - \left( \frac{K}{S} \right)_{Paper}  = c_{1} \left[\left( \frac{K}{S} \right)_{Pigment-1} - \left( \frac{K}{S} \right)_{Paper} \right] + .... + c_{n} \left[\left( \frac{K}{S} \right)_{Pigment-n} - \left( \frac{K}{S} \right)_{Paper} \right]
\end{equation}
We handpicked the pixel spectrum of the paper. Then we use equation \ref{eq:eq6} to formulate the unmixing process. However, few assumptions are made for the approximation of the K-M theory: (a) the pigment particles are smaller compared to the thickness of the layer, (b) diffuse illumination source and diffuse reflectance, (c) materials have high optical thickness, (d) the specular reflection is neglected. To ensure that pigment materials have high reflectance, Zhao et al \cite{zhao2008investigation} added appropriate amount of white paint to form a spectral database which they used for pigment identification using K-M theory. Liang et al \cite{liang2008pigment} showed that this unimixing method works even without adding any additional white pigments, where the pigment materials have low absorption coefficient and high scattering coefficient. However, note that we do not have a prior database of unique pigments or any information about the pigments present in our work.    

\section{Pigment Diversity Estimation and Unmixing}
\label{sec:pigmentdiversity}
In this section we will describe theoretical background on material diversity estimation in HSI using the Gram Matrix and MaxD techniques and spectral unmixing using the Non-Negative Least Square (NNLS) technique. 

In HSI, material diversity estimation implies the measurement of the dimensionality of the data. Several algorithms have been used to estimate the dimensionality of HSI datasets \cite{ziemann2010iterative,zhang2017endmember, gan2018endmember, parente2010survey}. We used the method introduced by Messinger et al \cite{messinger2012metrics} and used by Canham et al \cite{canham2011spatially} to estimate the materials diversity. Using the MaxD technique, $m$ number of endmembers are extracted \cite{schott2003subpixel}. The Gram Matrix $\mathbf{G}$ can be computed from a test set, $\left. \left\{ x_n\right\} \right\vert_{n=1}^{k}$, where the $\left\{ i,j\right\}^{th}$ element in the matrix is given by 
\begin{equation}
    \mathbf{G}_{i,j} = \langle x_i, x_j\rangle 
\end{equation}
which is the inner product between the $i^{th}$ and $j^{th}$ vectors in the test set. $\mathbf{G}$ is a $k\times k$ matrix where the number of vectors is $k$ in the test set. The determinant of the Gram Matrix is the square of the volume of the parallelotope generated by the test vectors. When the volume goes to $0$, the vectors within the set are no longer linearly independent. An independent set of enemembers implies a set of unique materials. Then using an iterative procedure the volume function can be plotted which indicates the dimensionality of the data which is shown in Figure \ref{fig:endmembers}a.

We used the NNLS method for pigment unmixing. The objective of unmixing is to find the abundance matrix given the HSI data and the endmember matrix. If the endmember matrix is $C$ and the data vector is $d$, NNLS tries to minimize the $\left \Vert C\cdot y - d \right\Vert_{2}^{2}$ loss function where $y \geq 0$. The matrix consisting $y$ for all the data points is called the abundance matrix.  

In this research we applied the unmixing in the K-M space .We transformed the endmember matrix and the HSI reflectance data to the K-M space and estimated the abundance matrix. Then we computed the difference matrix and augmented the abundance matrix with the difference matrix. If a vector in the abundance matrix is given as 
\begin{equation}\label{eq:eq8}
    \mathbf{x}_1 = 0.48\mathbf{E}_1 + 0.49\mathbf{E}_2 + 0.02 \mathbf{E}_3 
\end{equation}
the difference matrix is computed by taking the difference between the abundances of the constituent endmembers. For equation \ref{eq:eq8}, $(0.001, 0.46, 0.47)$ are the components of the difference matrix which is appended with the abundance matrix to make the feature descriptor more discriminative. Finally, we perform K-means clustering of the joint matrix and get our final class map. 

\section{Methodology}
\label{sec:method}
\begin{figure}[t]
\begin{center}
   \includegraphics[width = \textwidth]{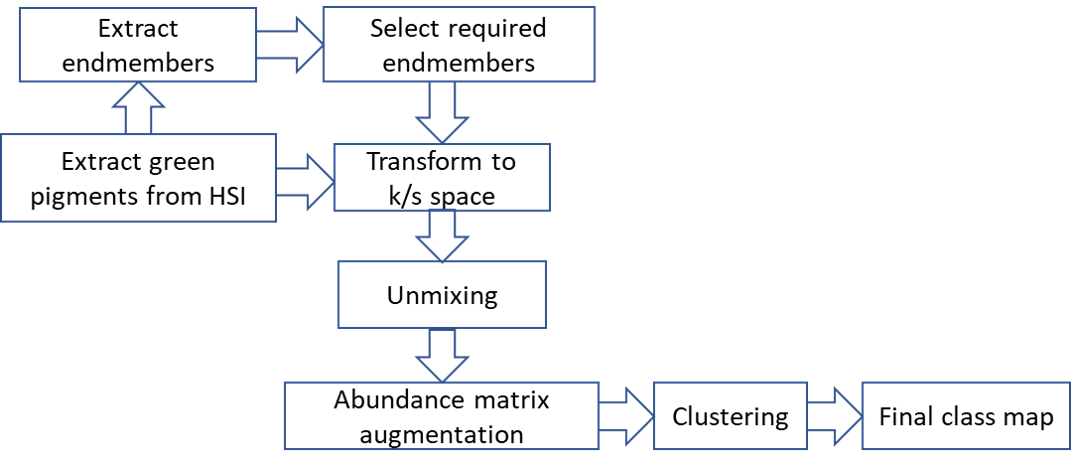}
\end{center}
   \caption{Workflow of our method.}
\label{fig:method}
\end{figure}

In this research, we estimated the pigment diversity of a predefined portion of the Selden Map using the K-M theory. The workflow of our method is shown in Figure \ref{fig:method}. The overall procedure is outlined below.

1. Cut specific portions of the `chips' and stitch them together to create a single patch.

2. Find the green pixels using Mahalanobis distance classifier (MDC).

3. Use the MaxD and Gram Matrix approach to estimate the dimensionality of the data and extract the endmembers.

4. Transform the data into the K-M space using the opaque form of single-constant K-M theory.

5. Use the NNLS technique to find the abundance matrix and augment the abundance matrix with the difference matrix.

6. Use K-means clustering to find the final classification map using the abundance plus difference matrix.

\section{Experiment}
\label{sec:experiment}
In this section we will explain the dataset, details of the methodology, evaluation metrics, and results and comparison. 

\subsection{Dataset}
\begin{figure}[t]
\begin{center}
   \includegraphics[width= \linewidth]{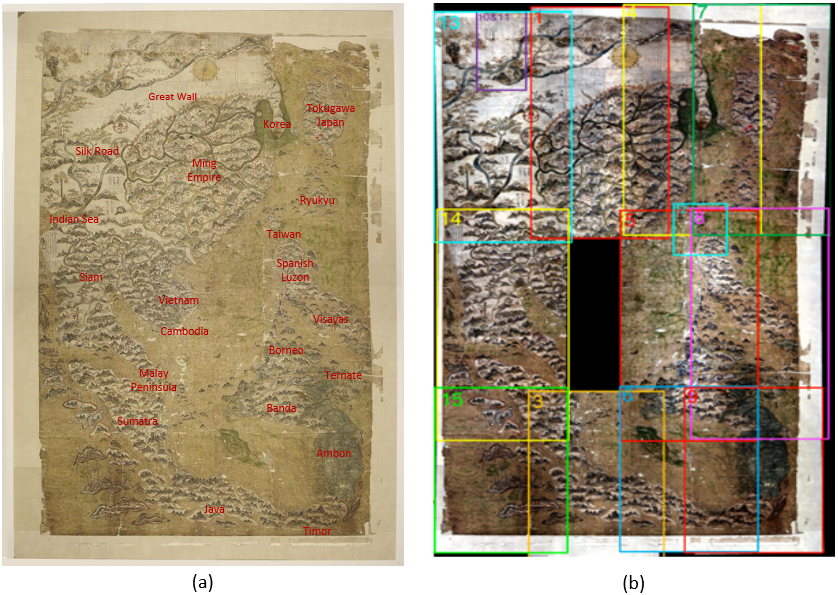}
\end{center}
   \caption{(a) RGB image of the Selden Map labelled with its major landmarks. (b) RGB rendering of the HSI `chips'. Notice that `chip' no. 2 is missing as it was lost during data transmission.}
\label{fig:fullMap}
\end{figure}

To capture high resolution HSI data of the overall map, the map was imaged in $12$ overlapping `chips'. All the overlapping `chips' are shown in Figure \ref{fig:fullMap}b after RGB rendering from the HSI data. The map data was captured using a hyperspectral imaging system. It is noted that the image data of `chip' 2 was lost during transmission. The HSI data has $334$ spectral bands ranging from $398.7$ nm to $1000.3$ nm where $\Delta \lambda = 1.8$ nm. The physical size of the Selden Map is 100cm x 160cm. In pixel space the overall HSI data dimension is $4784$ (pixels) x $7657$ (pixels) x $334$ (bands). So the spatial resolution is approximately $\Delta w \cong \Delta h \cong 0.0209$ cm/pixel.

The RGB image of the Selden Map with geographic location is shown in Figure \ref{fig:fullMap}a. Notice that the similar green pigments are used in different places of the map such as the Ming Empire, Korea, and Banda. 

\subsection{Experimental Details}
In this section we will discuss the overall experimental procedure to perform pigment diversity estimation.   
\subsubsection{Region of Interest selection}
\begin{figure}[t]
\begin{center}
   \includegraphics[width= \linewidth]{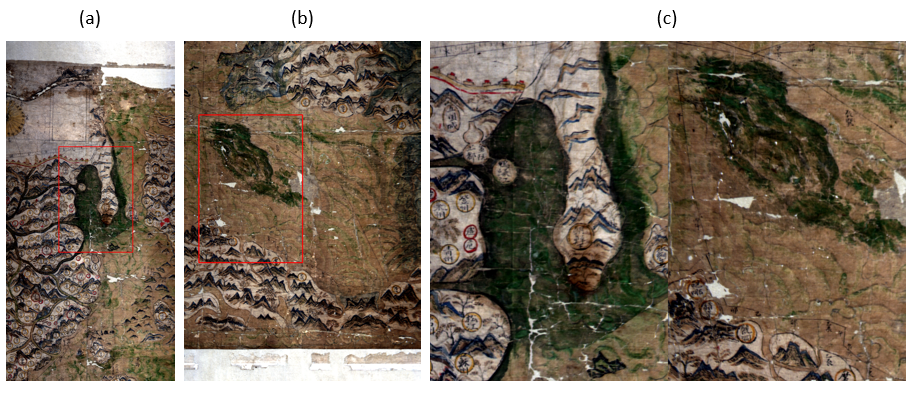}
\end{center}
   \caption{Target patch in (a) `chip' 4 and (b) `chip' 6. (c) The ROI is generated by horizontal placement of the two patches.}
\label{fig:roi}
\end{figure}

Our first step is to select the region of interest from the map. From the study of Bai et al\cite{bai2018} we found that Korea and the land between Java and Banda had large spectral angle error in the final classification map which motivated us to investigate further using a physics based approach for pigment unmixing. The cropped region of the RGB images from `chip' 4 and `chip' 6 are shown in Figure \ref{fig:roi}a and Figure \ref{fig:roi}b respectively. For computational simplicity, we stitch the two cropped regions side by side which is shown in Figure \ref{fig:roi}c. we call this patch as our region of interest (ROI). 

\begin{figure}[t]
\begin{center}
   \includegraphics[width= \linewidth, height = 8cm]{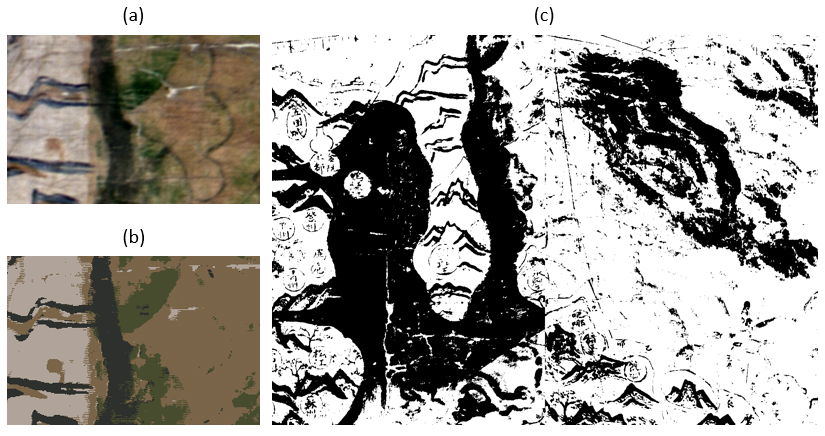}
\end{center}
   \caption{Segmentation of the ROI to extract green pixels. (a) A small patch is taken to train the Mahalanobis distance classifier. (b) K-means cluster map of the patch. (c) Final segmentation map as a binary image to extract the green pixels.}
\label{fig:segmentation}
\end{figure}

\subsubsection{Segmentation}
Our next step is to obtain the segmentation map to extract the green pigments. It is important to mention that the patch taken from `chip' 4 has some blackish pixels near the west border of Korea. We included those with the mountains in our segmentation map to extract the green pigments. We used both the RGB image and the HSI data to create an accurate segmentation map. We took a small patch from our ROI RGB image which is shown in Figure \ref{fig:segmentation}a. To get an estimate of the segmentation map we used the K-means clustering of the RGB image patch in RGB, Lab, and HSV colorspaces with and without augmenting the x and y coordinates of each pixel with the data. Clustering in the Lab colorspace with the x and y coordinates augmentation gave us the best possible estimate of the segmentation map which is shown in Figure \ref{fig:segmentation}b. Here we used 4 clusters and to show the clustering result, the color of each cluster is created by taking the mean of the red, green, and blue channels separately and replacing all the pixel values with the mean value within the same cluster. 
This segmentation mask is used as the training label of the corresponding HSI data to train the MDC. The final segmentation map is estimated by the prediction of the MDC from the overall HSI data corresponding to the ROI. 
The binary image of the segmentation map is shown in Figure \ref{fig:segmentation}c. The black pixels in the image represent the green and the dark pigments from the Selden map within the ROI. 

\begin{figure}[t]
\begin{center}
   \includegraphics[width= \linewidth, height = 6cm]{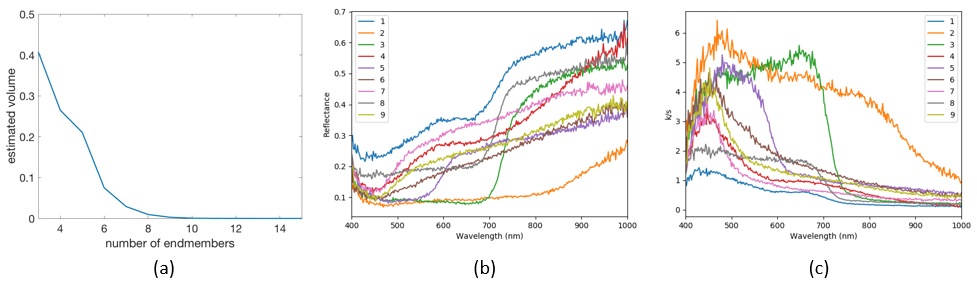}
\end{center}
   \caption{(a) Estimated volume function from Gram Matrix. (b) Extracted endmember spectra. (c) Endmember spectra in the K/S space.}
\label{fig:endmembers}
\end{figure}

\begin{figure}[t]
\begin{center}
   \includegraphics[]{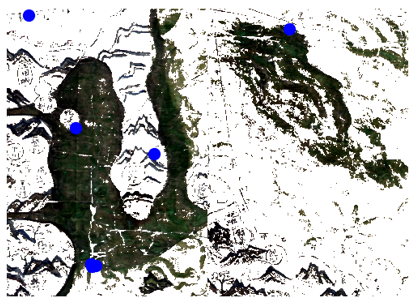}
\end{center}
   \caption{Endmember locations in the ROI image (blue dots). Two endmembers are not visibly distinguishable because they are spatially very close to their neighboring endmembers.}
\label{fig:endmemberloc}
\end{figure}

\subsubsection{Endmember Extraction}
Using the segmentation map we extracted the green pixels from the HSI for further analysis. Then we used the Gram Matrix technique to estimate the number of endmembers presented in the data identified by where the function approaches $0$, which is shown in Figure \ref{fig:endmembers}a. It is inferred from the figure that the number of endmembers present in the HSI is $9$. Then we used the MaxD technique to find those endmembers. The extracted endmembers are shown in Figure \ref{fig:endmembers}b which include both green pixels (bright around $550$ nm) and dark pixels (dark across the visible spectrum). The endmember locations in the original ROI image is shown in Figure \ref{fig:endmemberloc} with concentrated blue circles.

\begin{figure}[t]
\begin{center}
   \includegraphics[]{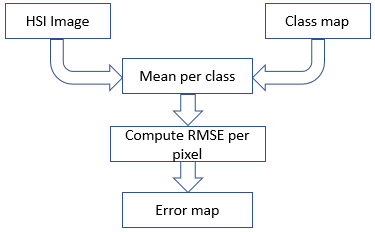}
\end{center}
   \caption{Workflow of computing the final classification error.}
\label{fig:error}
\end{figure}

\begin{figure}[t]
\begin{center}
   \includegraphics[width = \linewidth]{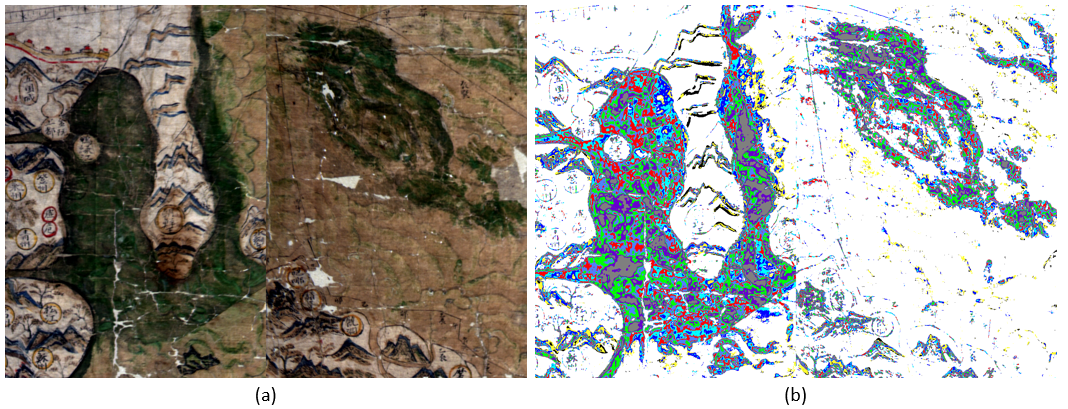}
\end{center}
   \caption{(a) RGB rendering of the ROI from HSI data. (b) The final classification map .}
\label{fig:finalclass}
\end{figure}

\subsubsection{K-M transformation}
Our next step is to transform the data and the endmembers into the K/S space. We used the single-constant opaque form of K-M theory given in Equation \ref{eq:eq3} for this transformation. The transformed data of the endmembers are shown in Figure \ref{fig:endmembers}c. 

\subsubsection{Unmixing and Clustering}
We unmix the HSI data in the K/S space using the NNLS technique. This technique enforces the abundances to be non-negative but the sum of the abundances for each pixel can be less or greater than $1$ where ideally it should be equal to $1$. This is to allow for within material variability and to avoid overfitting. This unmixing gave us the abundance matrix. We used the same technique used by Bai et al\cite{bai2017hyperspectral} which is described in section \ref{sec:pigmentdiversity} to compute the difference matrix from the abundance matrix. Then we augmented the abundance matrix with the difference matrix and used the K-means clustering algorithm to find the final classification map. 

\subsection{Metrics of Success}
For the success metrics we use the root mean squared error (RMSE) between the cluster mean and each of the pixels within that cluster. Our approach is to minimize the total RMSE for a certain class from the corresponding class mean. The workflow is shown in Figure \ref{fig:error}.

\subsection{Results}
\begin{figure}[t]
\begin{center}
   \includegraphics[width = \linewidth]{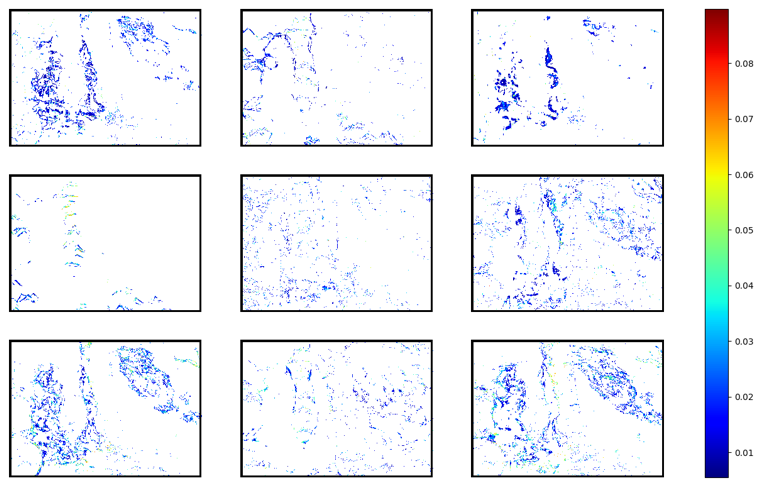}
\end{center}
   \caption{The spatial distribution of RMSE for different class means within the ROI.}
\label{fig:errordistribution}
\end{figure}

The final classification map comes from the clustering of the abundance map in the K/S space. We use the segmentation map to visualize the final classification result which is shown in Figure \ref{fig:finalclass}. From the classification result we can see that the boundary of Korea is drawn with a dark pigment which is not used in the part of the land between Banda and Java. We can also see that a darker green pigment is used significantly in Korea, but used marginally in the other region. We suspect that these changes were made during a modification of the map. However, few other green pigments are used in these two regions simultaneously. Our findings have some agreement with Bai et al \cite{bai2018}, where the author mentioned that similar pigments are used in these two places. However, our study provides deeper analysis of these places using K-M theory.    
The spatial distribution of the error map is shown in Figure \ref{fig:errordistribution}. Note that all the errors are small indicating a good model per pixel in the final classification map.

\section{Summary}
\label{sec:summary}
Kubelka-Munk (K-M) theory has been used to estimate pigment mixtures of modern paintings in hyperspectral imagery.  Here, K-M theory is utilized to classify pigments in the Selden Map of China, a navigational map from the early seventeenth century.  Hyperspectral data of the map was collected at the University of Oxford and can be used to estimate the pigment diversity, and spatial distribution, within the map. This work seeks to assess the utility of analyzing the data using K-M theory, as opposed to the traditional reflectance domain. From our study we found some similarities and dissimilarities of green pigments' spatial distribution between Korea and the land between Banda and Java. This method can be generalized to a tool to estimate pigment diversity in historical artifacts where little is known of the specific pigments and mixtures used. Geographers and cartographic historians may use this technique to understand the timeline of construction and modification of various historical artifacts.

\bibliography{report}   

\begin{thebibliography}{10}

\bibitem{bai2018}
Bai, D., Messinger, D.~W., and Howell, D., ``Pigment diversity estimation for
  hyperspectral images of the selden map of china,'' {\em Proc.SPIE}~{\bf
  10644},  10644 -- 10644 -- 17 (2018).

\bibitem{batchelor2013}
Batchelor, R., ``The selden map rediscovered: A chinese map of east asian
  shipping routes, c.1619,'' {\em Imago Mundi}~{\bf 65}(1),  37--63 (2013).

\bibitem{Kogou2016}
Kogou, S., Neate, S., Coveney, C., Miles, A., Boocock, D., Burgio, L., Cheung,
  C.~S., and Liang, H., ``The origins of the selden map of china: scientific
  analysis of the painting materials and techniques using a holistic
  approach,'' {\em Heritage Science}~{\bf 4},  28 (Sep 2016).

\bibitem{easton2011ten}
Easton, R., Christens-Barry, W.~A., and Knox, K.~T.,  [{\em Ten years of
  lessons from imaging of the Archimedes
  Palimpsest}{\nolinebreak\hspace{0.1em}]}, na (2011).

\bibitem{easton2010infinite}
Easton, R.~L. and Noel, W., ``Infinite possibilities: Ten years of study of the
  archimedes palimpsest,'' {\em Proceedings of the American Philosophical
  Society}~{\bf 154}(1),  50--76 (2010).

\bibitem{christens2011some}
Christens-Barry, W., Boydston, K., and Easton, R.,  [{\em Some properties of
  textual heritage materials of importance in spectral imaging
  projects}{\nolinebreak\hspace{0.1em}]}, na (2011).

\bibitem{easton2011some}
Easton, R., Knox, K., and Barry, W., ``Some properties of textual heritage
  materials of importance in spectral 27 imaging projects,'' in [{\em Proc. of
  the European Signal and Image Processing
  Conf.(EURASIP’11)}{\nolinebreak\hspace{0.1em}]},   1440--1444 (2011).

\bibitem{knox2011recovery}
Knox, K.~T., Easton, R.~L., Christens-Barry, W.~A., and Boydston, K.,
  ``Recovery of handwritten text from the diaries and papers of david
  livingstone,'' in [{\em Computer Vision and Image Analysis of Art
  II}{\nolinebreak\hspace{0.1em}]},   {\bf 7869},  786909, International
  Society for Optics and Photonics (2011).

\bibitem{easton2014statistical}
Easton, R.~L. and Kelbe, D., ``Statistical processing of spectral imagery to
  recover writings from erased or damaged manuscripts,'' {\em manuscript
  cultures}~{\bf 7},  35--46 (2014).

\bibitem{easton2003multispectral}
Easton~Jr, R.~L., Knox, K.~T., and Christens-Barry, W.~A., ``Multispectral
  imaging of the archimedes palimpsest,'' in [{\em
  null}{\nolinebreak\hspace{0.1em}]},   111, IEEE (2003).

\bibitem{easton2015rediscovering}
Easton, R.~L., Sacca, K., Heyworth, G., Boydston, K., Van~Duzer, C., and
  Phelps, M., ``Rediscovering text in the yale martellus map,'' in [{\em
  Information Forensics and Security (WIFS), 2015 IEEE International Workshop
  on}{\nolinebreak\hspace{0.1em}]},   1--6, IEEE (2015).

\bibitem{bai2017hyperspectral}
Bai, D., Messinger, D.~W., and Howell, D., ``Hyperspectral analysis of cultural
  heritage artifacts: pigment material diversity in the gough map of britain,''
  {\em Optical Engineering}~{\bf 56}(8),  081805 (2017).

\bibitem{bai2017pigment}
Bai, D., Messinger, D.~W., and Howell, D., ``A pigment analysis tool for
  hyperspectral images of cultural heritage artifacts,'' in [{\em Algorithms
  and Technologies for Multispectral, Hyperspectral, and Ultraspectral Imagery
  XXIII}{\nolinebreak\hspace{0.1em}]},   {\bf 10198},  101981A, International
  Society for Optics and Photonics (2017).

\bibitem{bowles2007optical}
Bowles, J.~H. and Gillis, D.~B., ``An optical real-time adaptive spectral
  identification system (orasis),'' {\em Hyperspectral Data Exploitation:
  Theory and Applications} ,  77--106 (2007).

\bibitem{zhang2005spectral}
Zhang, J., Rivard, B., and S{\'a}nchez-Azofeifa, A., ``Spectral unmixing of
  normalized reflectance data for the deconvolution of lichen and rock
  mixtures,'' {\em Remote Sensing of Environment}~{\bf 95}(1),  57--66 (2005).

\bibitem{liang2012advances}
Liang, H., ``Advances in multispectral and hyperspectral imaging for
  archaeology and art conservation,'' {\em Applied Physics A}~{\bf 106}(2),
  309--323 (2012).

\bibitem{kubelka1931contribution}
Kubelka, P. and Munk, F., ``A contribution to the optics of pigments,'' {\em Z.
  Tech. Phys}~{\bf 12},  593--599 (1931).

\bibitem{kubelka1948new}
Kubelka, P., ``New contributions to the optics of intensely light-scattering
  materials. part i,'' {\em Josa}~{\bf 38}(5),  448--457 (1948).

\bibitem{zhao2008investigation}
Zhao, Y., Berns, R.~S., Taplin, L.~A., and Coddington, J., ``An investigation
  of multispectral imaging for the mapping of pigments in paintings,'' in [{\em
  Computer image analysis in the study of art}{\nolinebreak\hspace{0.1em}]},
  {\bf 6810},  681007, International Society for Optics and Photonics (2008).

\bibitem{moghareh2017linear}
Moghareh~Abed, F. and Berns, R.~S., ``Linear modeling of modern artist paints
  using a modification of the opaque form of kubelka-munk turbid media
  theory,'' {\em Color Research \& Application}~{\bf 42}(3),  308--315 (2017).

\bibitem{liang2008pigment}
Liang, H., Keita, K., Peric, B., and Vajzovic, T., ``Pigment identification
  with optical coherence tomography and multispectral imaging,'' (2008).

\bibitem{duncan1940colour}
Duncan, D., ``The colour of pigment mixtures,'' {\em Proceedings of the
  Physical Society}~{\bf 52}(3),  390 (1940).

\bibitem{berns2002multiple}
Berns, R.~S., Krueger, J., and Swicklik, M., ``Multiple pigment selection for
  inpainting using visible reflectance spectrophotometry,'' {\em Studies in
  conservation}~{\bf 47}(1),  46--61 (2002).

\bibitem{guo1998application}
Guo, S. and Li, G., ``Application of kubelka-munk theory in device-independent
  color space error diffusion,'' in [{\em PICS}{\nolinebreak\hspace{0.1em}]},
  344--348 (1998).

\bibitem{ziemann2010iterative}
Ziemann, A.~K., Messinger, D.~W., and Basener, W.~F., ``Iterative convex hull
  volume estimation in hyperspectral imagery for change detection,'' in [{\em
  Algorithms and Technologies for Multispectral, Hyperspectral, and
  Ultraspectral Imagery XVI}{\nolinebreak\hspace{0.1em}]},   {\bf 7695},
  76951I, International Society for Optics and Photonics (2010).

\bibitem{zhang2017endmember}
Zhang, C., Qin, Q., Zhang, T., Sun, Y., and Chen, C., ``Endmember extraction
  from hyperspectral image based on discrete firefly algorithm (ee-dfa),'' {\em
  ISPRS Journal of Photogrammetry and Remote Sensing}~{\bf 126},  108--119
  (2017).

\bibitem{gan2018endmember}
Gan, Y., Hu, B., Liu, W., Wang, S., Zhang, G., Feng, X., and Wen, D.,
  ``Endmember extraction from hyperspectral imagery based on qr factorisation
  using givens rotations,'' {\em IET Image Processing}  (2018).

\bibitem{parente2010survey}
Parente, M. and Plaza, A., ``Survey of geometric and statistical unmixing
  algorithms for hyperspectral images,'' in [{\em 2010 2nd Workshop on
  Hyperspectral Image and Signal Processing: Evolution in Remote
  Sensing}{\nolinebreak\hspace{0.1em}]},   1--4, IEEE (2010).

\bibitem{messinger2012metrics}
Messinger, D.~W., Ziemann, A.~K., Schlamm, A., and Basener, W., ``Metrics of
  spectral image complexity with application to large area search,'' {\em
  Optical Engineering}~{\bf 51}(3),  036201 (2012).

\bibitem{canham2011spatially}
Canham, K., Schlamm, A., Ziemann, A., Basener, B., and Messinger, D.,
  ``Spatially adaptive hyperspectral unmixing,'' {\em IEEE Transactions on
  Geoscience and Remote Sensing}~{\bf 49}(11),  4248--4262 (2011).

\bibitem{schott2003subpixel}
Schott, J.~R., Lee, K., Raqueno, R., Hoffmann, G., and Healey, G., ``A subpixel
  target detection technique based on the invariance approach,'' in [{\em
  AVIRIS Airborne Geoscience workshop
  Proceedings}{\nolinebreak\hspace{0.1em}]},  (2003).

\end{thebibliography}
\bibliographystyle{spiebib}

\end{document}